# A hybrid deep-learning-metaheuristic framework for bi-level network design problems

Bahman Madadi[1*] and Gonçalo Homem de Almeida Correia[1]

1. Department of Transport & Planning, Faculty of Civil Engineering and Geosciences, Delft University of Technology, Delft, the Netherlands

*Corresponding author: b.madadi@tudelft.nl

Second author email: G.Correia@tudelft.nl

## ABSTRACT

This study proposes a hybrid deep-learning-metaheuristic framework with a bi-level architecture for road network design problems (NDPs). We train a graph neural network (GNN) to approximate the solution of the user equilibrium (UE) traffic assignment problem and use inferences made by the trained model to calculate fitness function evaluations of a genetic algorithm (GA) to approximate solutions for NDPs. Using three test networks, two NDP variants and an exact solver as benchmark, we show that on average, our proposed framework can provide solutions within 1.5% gap of the best results in less than 0.5% of the time used by the exact solution procedure. Our framework can be utilized within an expert system for infrastructure planning to determine the best infrastructure planning and management decisions under different scenarios. Given the flexibility of the framework, it can easily be adapted to many other decision problems that can be modeled as bi-level problems on graphs. Moreover, we foreseen interesting future research directions, thus we also put forward a brief research agenda for this topic. The key observation from our research that can shape future research is that the fitness function evaluation time using the inferences made by the GNN model was in the order of milliseconds, which points to an opportunity and a need for novel heuristics that 1) can cope well with noisy fitness function values provided by deep learning models, and 2) can use the significantly enlarged efficiency of the evaluation step to explore the search space effectively (rather than efficiently). This opens a new avenue for a modern class of metaheuristics that are crafted for use with AI-powered predictors.

# 1    Introduction

Solving bi-level and NP-hard combinatorial optimization problems in the transport domain is often very challenging [1]. Traditional approaches of tackling such problems include exact solutions, approximation algorithms, and heuristics. Exact solutions might be computationally impossible to obtain for large instances or non-existent for some problem variants. Approximate algorithms are often computationally more advantageous but suffer from weak optimality guarantees and are not always available. Heuristics are usually the fastest, yet they lack theoretical foundation and often require substantial domain knowledge of the problem and a trial-and-error process for successful implementations [2], which can yield quite variable results.

On the other hand, many real-world applications require solving the same problem or a subproblem within the main problem with slightly different input regularly (e.g., shippers and freight forwarders deal with vehicle routing problems on a daily basis). None of the traditional approaches exploits this property of recurrence in optimization problems. In addition, many optimization problems are defined on graph-structured data, yet many of the solution methods do not sufficiently explore the graphs' structure to accelerate the solving process.

This provides an opportunity for training models that can learn to approximate such problems, particularly when the problems are defined on graph-structured data. The main motivation behind this approach is that after a one-time effort to train a model (e.g., a multilayer perceptron), this model can be used numerous times to approximate many instances of the problem within a small fraction of the time required for formal methods to solve new variations.

Moreover, artificial neural networks defined on graphs, which are often referred to as graph neural networks (GNNs), can discover patterns and structures in graphs using their deep architecture that engineers (of heuristics) and formal optimization methods may not be able to discover. In this paradigm, the deep learning model (e.g., a GNN) becomes the engineer of the algorithm and the researcher's efforts shift from engineering algorithms to designing deep learning models that can themselves engineer solution algorithms by parametrizing decision rules and policies that can lead to powerful algorithms [3].

The success of deep learning in various fields has prompted researchers to use them to train models and heuristics for tackling recurrent optimization problems [2]–[7]. The prospect of designing new heuristics without the need for having profound knowledge of the specific domain under consideration, and hand-crafting details of the algorithms, makes deep learning an even more promising approach for tackling NP-hard problems [5].

However, so far, efforts have mainly focused on classical combinatorial optimization problems, such as the traveling salesman problem (TSP) and the vehicle routing problem (VRP) [6], [8]. More challenging problems such as bi-level optimization problems, particularly the ones in the transportation domain, e.g., network design problems [9], have not received much attention from deep learning researchers yet. Even though bi-level deep learning architectures [10] and bi-level modeling of single-level graph optimization problems [11] have been proposed in the literature, deep learning methods for solving bi-level programing problems have not been proposed yet.

Although some bi-level programming problems might be too challenging to be solved using end-to-end deep learning methods, hybrid deep learning approaches have been shown to perform well on complex optimization problems where the optimization procedure can be decomposed to delegate some tasks to (deep) machine learning models and some others to classical approaches [8]. For instance, supervised learning (decision trees in particular) has been used within branch-and-bound methods to select branching variables [12]. Bagloee et al. [13] have used a combination of linear regression as a supervised learning model and an exact solver to deal with the upper-level and the lower-level of bi-level programming problems, respectively. Another example of such hybrid methods is combining unsupervised learning with simulation-based optimization to learn the best combination of metrics to use within the simulation to make optimal planning decisions in expert systems [14]. The prime advantage of such hybrid methods is their flexibility and explainability.

The main aim of this study is to demonstrate the potential of deep learning for tackling network design problems (NDPs) as an essential category of bi-level programming problems, which has not been done before. In transport literature, strategic decisions regarding modifications to road networks (e.g., adding or removing lanes or stretches of roads) are considered within the well-known road NDP framework [9], [15]. NDPs are often modeled as Stackelberg leader-follower games, which are usually bi-level programming problems where the leader (upper level) decides on the infrastructure (road network) and the followers react with their travel choices, namely their routes within the network (lower level). The objective of the leader (i.e., transport planner) is typically to maximize the social benefits provided to the followers (e.g., reduction in total system travel time) given a certain budget for the infrastructure adjustment cost (e.g., adding lanes), whilst the objective of the followers (i.e., travelers) is to minimize their individual travel time (which do not necessarily lead to minimum total system travel time) by selecting routes, which leads to the user equilibrium (UE) traffic assignment problem [16].

Therefore, in this study, we show how to learn to approximate the UE traffic assignment problem, i.e., the lower-level problem of NDPs, using a GNN and propose a hybrid deep-learning-metaheuristics approach using the proposed GNN and the genetic algorithm (GA) to deal with a class of bi-level NDPs, namely road network design problems, which are commonly used to support intelligent road network infrastructure planning decisions. We benchmark the performance of our framework against an exact solution method, a mathematical program implemented in a solver, using extensive numerical studies on three road networks. To the best of our knowledge, this is the first study that tackles bi-level NDPs using deep learning on graphs. We believe our proposed framework can be applied to many other NDP variations as well and eventually become a common decision support tool for road planners.

The specific contributions of this study are as follows:

- Combining the knowledge of machine learning and operations research with specific knowledge of the transportation field to tackle a challenging and practical transportation problem, namely NDP, which has not received attention from machine learning scientists yet.
- Proposing a novel hybrid solution framework for tackling bi-level NDPs, which is computationally super-fast, particularly the idea of using the inferences made by a trained GNN for fitness function evaluations of metaheuristics within a bi-level framework is novel.
- Showcasing the potential and scalability of the proposed framework by conducting extensive numerical experiments including three road networks, two network design problems, six variants for each problem on each network (36 problem variants in total), attempting to solve each instance with an exact solution procedure as benchmark, performing multiple replications of our solution procedure, and providing a statistical analysis of the results.
- Paving the way for future research on using deep learning to solve NDPs by creating standard benchmark problem variants and datasets that can be used for performance comparisons in future studies.
- Facilitating future research on this topic by providing a research agenda that identifies opportunities for new developments in this area.

The structure of this article is as follows. Section 2 provides a background on UE traffic assignment problems, NDPs, and GNNs. Section 3 describes our proposed hybrid deep-learning-metaheuristics framework. Section 4 presents numerical experiments using three case studies to benchmark the performance of our framework against an exact solution procedure. Section 5 provides a discussion on the results. And the last section includes conclusions and future research directions.

## 2  Network design problem description and preliminaries on graph neural networks

As mentioned earlier, our framework deals with road NDPs using a hybrid method based on a GNN. Therefore, we provide a brief background for road NDPs and GNNs in this section.

### 2.1  Network design problem

As briefly mentioned in the introduction, road network design problems include the UE traffic assignment problem at the lower level and the design problem at the upper level. The following subsections elaborately describe each subproblem.

#### 2.1.1  The lower-level user equilibrium (UE) problem

Consider a road network represented by the graph $G = (V, E)$ where $V$ is the set of nodes, $E$ denotes the set of edges (i.e., links or road segments), and an adjacency matrix of size $|V| \times |V|$ representing which nodes are connected by an edge. The UE problem includes assigning the origin-destination (OD) flows of travelers between the nodes through edges in the network in such a way that each traveler cannot reduce their travel time by changing routes. In its classical form, UE is a convex non-linear mathematical programming problem formulated as follows.

**UE:**

$$\min \quad \sum_{(i,j) \in E} \int_0^{x_{ij}} t_{ij}(x_{ij}) \tag{1}$$

$$s.t. \quad \sum_{j \in V; (i,j) \in E} x_{ijs} - \sum_{j \in V; (j,i) \in E} x_{jis} = d_{is}, \quad \forall i \in V, \forall s \in D \tag{2}$$

$$\sum_{s \in D} x_{ijs} = x_{ij}, \quad \forall (i,j) \in E \tag{3}$$

$$x_{ijs} \geq 0, \quad \forall (i,j) \in E, \forall s \in D, \tag{4}$$

where $t_{ij}$ is the travel time of link (edge) $(i,j)$ from node $i$ to node $j$, $D$ is the set of destination nodes ($D \subseteq V$) and $d_{is}$ is the demand from node $i$ to destination node $s$. $x_{ijs}$ is the flow traveling to destination $s$ on edge $(i,j)$ and $x_{ij}$ is the total flow on edge $(i,j)$. It can be demonstrated that equation (1) minimizes the individual travel time for each traveler, equation (2) guarantees demand satisfaction for each OD, equation (3) ensures flow conservation, and equation (4) prevents negative flows. The edge travel time is usually estimated via the Bureau of Public Roads (BPR) function:

$$t_{ij} = t_{ij}^0 (1 + \alpha (\frac{x_{ij}}{\pi_{ij}})^\beta)$$

where $t_{ij}^0$ is the free flow travel time, $\pi_{ij}$ is the edge capacity and $\alpha, \beta$ are parameters. This makes the BPR function a polynomial function, rendering the UE problem a non-linear but still convex mathematical programming problem. It is worth noticing that other formulations exist for the UE problem. However, the above-mentioned formulation stands out as the most suitable one for using general-purpose solvers to obtain a solution for the problem.

2.1.2 The upper-level design problem

In the upper-level design problem, the objective is to minimize specific criteria, usually total travel time, by making network modifications (e.g., adding new lanes) given a predefined budget constraint while evaluating the lower-level problem at its equilibrium state. The following formally represents the classical discrete NDP.

**NDP:**

$$\min \sum_{(i,j) \in E} t_{ij}(x_{ij}) \tag{5}$$

$$\text{s.t.} \sum_{(i,j) \in E_1} y_{ij} c_{ij} \leq B \tag{6}$$

$$x_{ij} \in \arg\min \quad UE \tag{7}$$

$$y_{ij} \in \{0,1\}, \quad \forall (i,j) \in E_1, \tag{8}$$

where $y_{ij}$ is a binary decision vector assuming the value of one when a new lane is added to the edge $(i,j)$ and zero otherwise, $c_{ij}$ denotes the cost of changing/upgrading the edge $(i,j)$, (e.g. adding a lane or repaving the road), and $B$ is the total budget. Equation (7) implies flows used to calculate total travel time must be equilibrium flows obtained by solving the UE.

Different variants of NDP usually include one or more set of constraints to relate the edge travel time to the modifications made on the network (e.g., capacity addition due to new lanes). We will discuss two of these variants below, which are used in the numerical experiments in this paper.

2.1.3 Network design problem with lane additions (NDP-LA)

This is one of the classical discrete NDPs where the road network can be improved by adding new lanes to edges given a certain budget constraint. The mathematical formulation of the NDP-LA will include equations (1)-(8) as well as the equation (9) below to define the capacity of each edge based on the value of the upper-level decision variables $y_{ij}$, its initial capacity $\pi_{ij}^0$ and the capacity of an extra lane $\pi_{ij}^0$.

$$\pi_{ij} = \pi_{ij}^0 + y_{ij}\pi_{ij}^1, \quad \forall (i,j) \in E_1, \tag{9}$$

2.1.4 Network design problem with lane swap (NDP-LS)

This problem occurs when there is asymmetric demand in a certain period (e.g., morning peak hour), therefore, the traffic manager dynamically assigns lanes from one direction of an edge to the other direction to account for the asymmetric traffic. This is a problem that occurs daily in many cities around the world. Mathematically, NDP-LS is formulated using equations (1)-(8), excluding the budget constraint in equation (6), whilst equations (10)-(11) below are added to ensure the changes imposed by the lane swaps, namely in terms of capacity and unique existence of one of the traffic directions in the newly created lane.

$$\pi_{ij} = \pi_{ij}^0 + y_{ij}\pi_{ij}^1, \quad \pi_{ji} = \pi_{ji}^0 - y_{ij}\pi_{ji}^1, \quad \forall (i,j) \in E_1, \tag{10}$$

$$y_{ij} + y_{ji} \leq 1, \quad \forall (i,j) \in E_1, \tag{11}$$

## 2.2 GNN

The application of neural networks (NNs) for optimizing decisions in combinatorial optimization problems dates back to the 1980s when [17] applied a what came to be called a Hopfield-network for solving small TSP instances. However, as explained before, NNs are not per se aware of the structural properties of the graphs representing certain combinatorial optimization problems such as TSP and VRP. GNNs on the other hand are defined on graphs that characterize specific problems (e.g., TSP, citation networks, and social network analysis) and have shown strong representation power in dealing with problems defined on graphs [18]. This is the main reason why they have enjoyed great success in dealing with graph-related problems in recent years [19]–[24].

In general terms, GNNs use a neighborhood aggregation scheme (also known as message passing) where each node of the graph receives and aggregates feature information (messages) from neighboring nodes to calculate its new state in an iterative manner and use the final state of the nodes after certain number of iterations to make predictions regarding nodes, edges or the entire graph [25], [26].

To formally define GNNs, consider a graph $G = (V, E)$ where $V$ represents the set of nodes (vertices) each having the node feature vector $x_v$ for $v \in V$, $E$ denotes the set of edges (links) with edge feature vectors $x_e$, and an adjacency matrix of size $|V| \times |V|$ represents which nodes are connected by an edge. A GNN uses the node features $x_v$, (optionally) the edge feature $x_e$ and the graph topology captured in the adjacency matrix to a learn a representation or state $h_v$ for each node $v$. The node state is calculated using a neighborhood aggregation schemes where the state of the node is iteratively updated based on the state of its neighboring nodes. After $k$ iterations of aggregation, a node's state captures the structural information within its $k$-hop network neighborhood. Formally, the $k$-th layer of a GNN is calculated as:

$$a_v^{(k)} = \text{AGGREGATE}^{(k)}(\{h_u^{(k-1)} : u \in \eta(v)\}), \tag{12}$$

$$h_v^{(k)} = \text{COMBINE}^{(k)}(h_v^{(k-1)}, a_v^{(k)}), \tag{13}$$

where $h_v^{(k)}$ is the feature vector of node $v$ at the $k$-th iteration/layer, $h_v^{(0)} = x_v$, and $\eta(v)$ is a set of nodes adjacent to $v$ (i.e., neighborhood of the node). The choice of $\text{AGGREGATE}^{(k)}$ and $\text{COMBINE}^{(k)}$ defines the GNN variant and has a major influence on its capabilities.

For instance, in the pooling variant of GraphSAGE [27], the following aggregation is used:

$$a_v^{(k)} = \text{MAX}(\{\text{ReLU}(W . h_u^{(k-1)}), \forall u \in \eta(v)\}), \tag{14}$$

where W is a learnable matrix, and MAX represents an element-wise max-pooling. The COMBINE step here is a concatenation of the current aggregate score and the previous iteration's state.

A more sophisticated node embedding scheme uses a multilayer perceptron (MLP) to aggregate neighborhood information. This is utilized in the graph isomorphism network (GIN) introduced in [18] where the authors use the graph isomorphism test to argue for the representation power of GIN using MLPs. GIN updates the node state as:

$$h_v^{(k)} = \text{MLP}^{(k)}((1 + \varepsilon^{(k)}) . h_v^{(k-1)} + \sum_{\forall u \in N(v)} h_u^{(k-1)}), \tag{15}$$

where the neighborhood aggregation function $h_v^{(k)}$ is an MPL rather than a straightforward function (e.g., max, mean, sum) and $\varepsilon$ is a learnable parameter.

In this study, we use GIN with a few modifications as our preferred GNN in the hybrid deep-learning-metaheuristics framework. The modifications will be discussed in the next section where we introduce our framework.

## 3    The hybrid deep-learning-metaheuristic framework (GIN-GA)

Simply put, our framework includes training a GIN to approximate the UE and using it in combination with a genetic algorithm (GA) to approximate the solution of NDPs. Given the bi-level structure of the NDP, the GA attempts to find optimal or near-optimal solutions for the upper-level problem while using GIN to approximate the lower-level problem (i.e., the UE) in its fitness evaluations. Deploying such a framework requires some preparation steps, which are described in this section. Note that our framework is very flexible and allows using any deep learning model to learn to approximate the UE and any heuristic to find solutions for the NDP. In this study, we use GIN as our preferred deep learning model and GA as our choice of metaheuristics, but as discussed in the conclusion section, many other deep learning models and heuristic algorithms could fit this framework. Figure 1 presents a visual summary of the steps involved with our hybrid framework. In the following subsections of this section, which correspond to the grey boxes in Figure 1, we describe each step in detail.

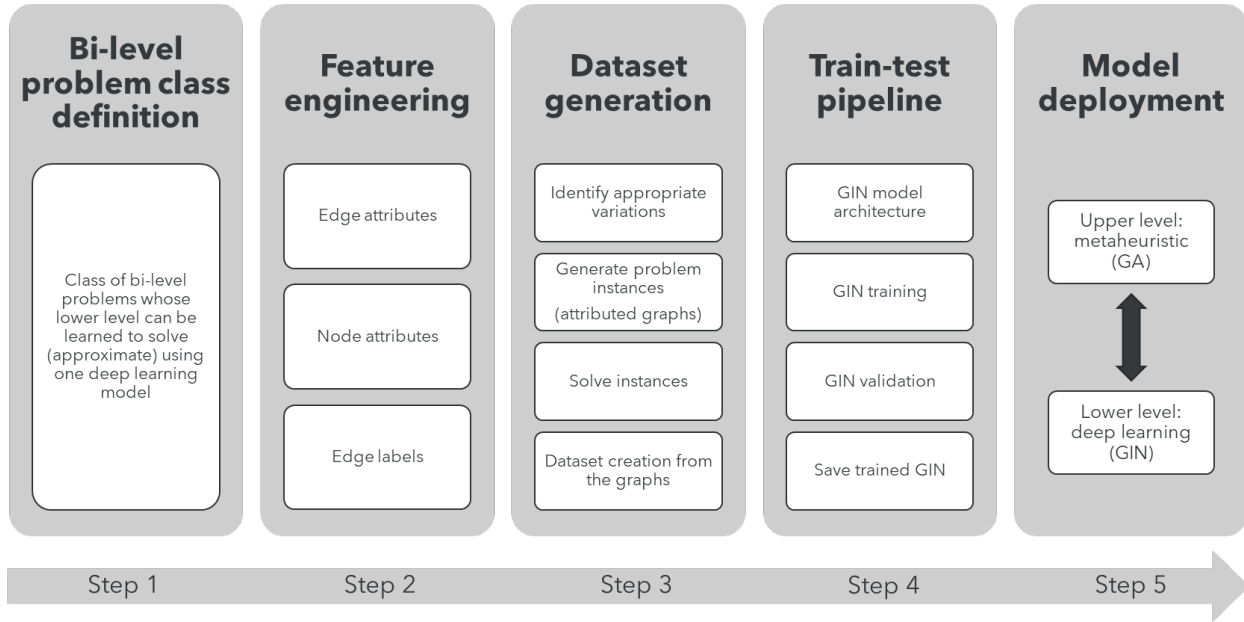

*Figure 1 Summary of the steps of hybrid deep-learning-metaheuristic framework (GIN-GA) for bi-level NDPs*

### 3.1 Bi-level problem class definition

The framework introduced in this article is suitable for a class of bi-level problems whose lower level can be solved using a single algorithm and thereby can be learned to be approximated using a single deep learning model. In this study, we focus on NDPs with upper-level decision variables indicating changes in edge capacity and the lower level (UE) problems including capacity and demand variations. The two NDP variants introduced earlier, namely, NDP-LS and NDP-LA, meet these criteria and belong to this class of NDPs. Therefore, we use these two problems in our numerical experiments. However, many other NDPs belong to this category [9] and can be tackled using this framework.

### 3.2 Feature engineering

Most solution procedures for solving NDPs entail solving the UE problem numerous times, which justifies the one-time effort of training a model to approximate the UE given that after this one-time training, the deep learning model (in this case the GIN) can find high-quality solutions to the UE problem in an extremely small fraction of the time required for solvers to solve this problem.

In order to train the GIN model to solve instances of the UE, first we need to define a graph representation in such a manner that each UE problem instance can be uniquely defined by an attributed graph including edges, nodes, edge attributes, and node attributes. Moreover, the solution to each UE instance needs to be represented by a vector of edge labels. Therefore, each data point (graph) used for training includes an edge label vector representing the equilibrium flows (i.e., the solution to the UE problem), and an attributed graph, which includes a road network and a travel demand matrix. For each edge of the graph, edge features include capacity and free flow travel time, and for each node, node features include incoming and outgoing demand from and to all other nodes. Therefore, the edge features are $|E| \times 2$ matrices, node features are $|V| \times |V|$ matrices and edge labels are $|V| \times 1$ vectors.

### 3.3 Dataset generation

The first step in generating a training dataset is identifying relevant problem instances. We want GIN to be able to approximate solutions to any UE problem with capacity and demand variations. Therefore, we first generate variations of UE with perturbations on demand and capacity. Then, we solve each problem instance using an exact solver, and save equilibrium flows as edge labels. This process is elaborated on in Section 4. Finally, we create a dataset including all graphs (i.e., data points) with edge and node attributes, representing UE problem instances, and one vector of edge labels (UE flows) for each graph.

### 3.4 Train-test pipeline

After creating a dataset with solved UE instances, we divide the dataset into training, validation, and test sets, and then train and validate the GIN to predict the equilibrium UE flows (i.e., edge labels) as an edge regression task. Regarding the model architecture, the main components of GIN were introduced in the previous section and are described in detail in [18]. The key modifications made in this study to tailor GIN to solving the UE are listed below.

- For neighborhood aggregation, instead of the standard adjacency matrix (with zeros and ones), a weighted version is used where instead of ones representing the existence of edges between two nodes, the value of free flow travel time divided by the edge capacity is used to capture the impacts of different edge capacities, speeds, and lengths.
- Edge scores (the values predicted by the model for UE flows on each edge) are based on an MLP that takes the node features of the source and sink node as well as the edge features of the corresponding edge as input and returns a score for UE flow on the edge.
- The model is trained with the mean square error (MSE) difference between edge labels and edge scores as the loss function, yet the value we are interested in is the total travel time, which is calculated afterwards using the BPR function. This is essentially equivalent to performing a graph regression task (as opposed to edge regression, which is used here) with the BPR function as the readout or graph aggregation function. However, since based on the knowledge of the problem we know the best readout function in this case is the BPR function, we use it instead of experimenting with different functions.

## 3.5   Model deployment

The trained GIN is deployed within a hybrid framework in combination with a GA to tackle the NDP in an iterative optimization-assignment scheme, where the GA deals with the upper-level (optimization) problem and the GIN approximates the lower-level UE (traffic assignment) problem. The decision variables of the upper-level correspond to the edges on which the lane modifications happen, and the objective function is the total travel time (equation (5)). The GA fitness evaluations are performed using the inferences made by the trained GIN, which predict UE edge flows for calculating the objective function. For the NDP-LA variants where there is a budget constraint, a penalty term is added to the objective function value to ensure budget-feasible solutions by penalizing the infeasible solutions. Since GNNs can make inferences for an entire batch very efficiently, the inferences are made in batches of size equal to the population size of the GA. It will be shown in the next section that such deployment of trained GNN model inferences can reduce the GA fitness evaluation times by a factor of at least 1000. Upon termination of the GA, a certain number of the solutions within the population of the last generation is evaluated using an exact UE solver to identify the best-found solution based on the exact objective function value. The number of the solutions of the last generation devalued is a model parameter. We will refer to the proposed framework including the GIN and the GA as GIN-GA in the remainder of this article.

## 4   Numerical experiments

### 4.1   Experiment setup

To showcase the performance of our proposed framework, we conduct extensive numerical experiments on three road networks, two network design problems, and six variants for each problem on each network, which culminates in 36 problem variants in total. The road networks studied are Sioux Falls, Eastern Massachusetts, and Anaheim, which are selected from the well-known "transportation networks for research" repository [28] to provide a small, medium and large network, respectively, for benchmarking purposes. Table 1 provides a summary of their characteristics, and they will be further described in dedicated subsections. The NDPs considered are NDP-LA and NDP-LS, described in Section 2.1. Variants of NDP-LA are created by perturbing the budget and number of lanes on each edge (which defines how much capacity is added to the edge by adding a lane). Variants of NDP-LS are created by perturbing feasible edges for lane swap, number of lanes on each edge, and demand. Demand variation matters in this case since the daily lane-swap plan is affected by the day-to-day variations in demand. The demand for each OD pair on each variant is generated by uniform perturbations within 20% of the original demand.

*Table 1  Main characteristics of the networks used in case studies*

| Case study | Number of Edges | Number of nodes | Number of OD pairs |
|---|---|---|---|
| **Case study 1: Sioux Falls** | 76 | 24 | 576 |
| **Case study 2: Eastern-Massachusetts** | 258 | 74 | 5476 |
| **Case study 3: Anaheim** | 914 | 416 | 1444 |

## 4.2 Benchmarking

To benchmark the performance of our framework, we use an exact solution procedure, namely, the system-optimal relaxation-based (SORB) method [29] with piece-wise linear approximations of the BPR travel time function. The SORB method takes advantage of the fact that an optimal solution for an NDP under the System Optimal (SO) principle can serve as a good approximation for a solution under the UE principle. The method sorts the solutions of relaxed subproblems (i.e., SO-relaxation-based solutions) in ascending order based on their objective function value (i.e., total travel time). The optimality of the solution is ensured when the lower bound of the total travel time for unexplored solutions under the UE principle is greater than or equal to the total travel time of a known solution under the UE principle. For details, the reader is referred to [29].

SORB is an exact solution procedure, which means that given sufficient computation time and memory for each instance of the problem, this method is guaranteed to find the global optimum, which is the ultimate measure for benchmarking any heuristic solution procedure for optimization problems. However, it is known that NDPs are among the most challenging problems in the transport domain and exact solutions for such problems are rare and cumbersome to implement [9], [15], hence even medium-size NDPs can present serious computational challenges [30] and become practically infeasible due to the need for solving many subproblems, each one being a large mixed-integer non-linear programming problem. Fortunately, it has been shown that the solution found by the SORB within the first 10 iterations is often optimal or within a very narrow gap of the lower bound for the optimal value [29], [30], even though the method requires many iterations to confirm the optimality of the found solution by iteratively tightening the lower bound.

We choose therefore to run the SORB method with a four-hour time limit for each problem instance as a strong heuristic for benchmarking the performance of our framework. Since the aim of this study is to show that our proposed framework can find high-quality and optimized (not necessarily optimal) solutions in a very short time, we run GIN-GA with a one-minute time limit for each problem instance and compare the solutions to the solutions found by SORB in four hours, i.e., the computation time limit for SORB is 240 times the computation time limit of GIN-GA. Since GIN-GA is stochastic, for each problem instance, we run five replications of GIN-GA and provide a statistical analysis of the results to show the consistency of its performance as well. To demonstrate its scalability, we use three different road networks with gradually increasing sizes and analyze its performance on networks with different sizes as well.

The following summarizes the procedure for performance comparisons used in this study.

o   For each network (Sioux Falls, Eastern Massachusetts, Anaheim):
    o   For each problem (NDP-LA & NDP-LS):
        o   For each variant (feasible lane and lane capacity variations):
            - Run SORB with four-hour time limit & record key performance indicators (KPIs).
            - Run GINGA with one-minute time limit five times & record KPIs.
            - Compare and statistically analyze minimum, maximum, average and confidence intervals for KPIs.

### 4.3 Key Performance Indicators (KPIs)

When approximating the UE with GIN, our main variable of interest is total travel time. However, our experiments showed that using edge flows as labels to train the GIN yields better results for predicting total travel times. Therefore, we used edge flows as labels and calculated total travel times using the BPR function. It should be noted that this is equivalent to using the BPR function as the readout function (or graph aggregation function) in a graph regression task using GIN. As for the loss function, we used mean squared error (MSE) between edge labels and model inferences (predictions) for edge flow values. Regarding model accuracy, we calculate and report total travel time prediction accuracy (mean absolute percentage (MAP)). To show how well the trained model fits the data, we measure and report R-squared.

As for the NDPs, the main performance indicator was total travel time (TTT) as calculated in equation (5). Therefore, we calculate the TTT (i.e., the NDP objective function value) obtained by SORB for each problem instance as well as the minimum, average, and maximum TTT across five stochastic replications of GIN-GA runs. In addition, we calculate the minimum, average, maximum, and 95% confidence intervals for the gap between the TTT obtained by SORB and the TTTs obtained by GIN-GA (i.e., the TTT gap) across five runs for each problem variant.

### 4.4 Hardware and software

SORB and GIN-GA method used in experiments presented in this study were coded in Python and ran on a Windows PC with an Intel(R) Core(TM) i7-1185G7 CPU @ 3.00 GHz and 16 GB RAM. No GPU was used for the experiments reported in this study.

To prepare the datasets for training GINs, we utilized AequilibraE [31], an open-source and comprehensive Python package for transportation modeling to solve the UE problems using the bi-conjugate Frank-Wolfe (BFW) algorithm [32], which is shown to provide the best performance among the existing algorithms for solving the UE traffic assignment problems [32]. The computation times of solving UE problem instances using AequilibraE based on the BFW algorithm to reach the gap threshold of 10e-6 for case studies considered here given the hardware described above is always below ten seconds. Note that convex (non-linear) UE problems such as the ones considered in this study can be solved using exact mixed integer non-linear programing solvers as well. However, for large networks, computation times of such solvers are practically infeasible when numerous instances of UE problems need to be solved.

To solve the NDPs with SORB, we coded SORB in Python using the CPLEX solver [33], one of the best-in-class optimization solvers, to solve the mixed-integer non-linear subproblems (with piece-wise linear approximations) within iterations of SORB. The UE subproblems within SORB were solved by means of AequilibraE using the BFW algorithm, which provides significant computational gains (approximately by a factor of 10) compared to solving the UE with general solvers as in the original study proposing the SORB method.

The graph features and datasets for training GIN were processed using the DGL package in Python for more efficiency, GIN was coded using the PyTorch package in Python, and its hyper-parameter tuning was performed by means of Bayesian optimization using the Hyperopt package [34] in Python with maximum 100 samples per network. The GA was implemented using the Geneticalgorithm2 package [35] in Python in combination with inferences made by the trained GIN models using PyTorch. Adam optimizer [36] was used in training to optimize the GIN. The training requires a one-time effort of 4-16 hours (depending on the network size, dataset size, and the number of epochs) on a regular desktop computer, described above (which could be accelerated using GPU).

**4.5  Case studies**

4.5.1  Case study 1: Sioux Falls Network

The first (and the smallest) case study is based on the Sioux Falls network, which is popular among transport researchers for benchmark studies. The network data is available at [28] and includes 76 edges, 24 nodes and 576 OD pairs. For the Sioux Falls network, we created a dataset with 20,000 solved instances of UE with perturbations in demand and edge capacity. 18,000 of these instances were used for training, 1,000 for validation, and 1,000 for testing. Note that validation and testing in this context are not important since the ultimate test of the model in this framework is how well it performs for inference (in combination with GA) during the optimization process, which is discussed below.

After hyper-parameter tuning, the best training accuracy and test accuracy (MAP) achieved for this network were 97.30% and 96.51%, respectively. The Best R-squared for training and test were 98.91% and 97.93% respectively, which indicates strong expandability of the variations in input by the model. The trained model is used for inferences in all variants and scenarios reported in this case study. Figure 2 summarizes the training results for the Sioux Falls network.

It should be noted that being able to deal with different demand profiles at least within a certain range of the known demand is crucial for generalization capacity and applicability of this method. Without variations in demand, higher accuracies could be achieved with GIN. However, that would mean that for every variant of the problem or with each small perturbation in demand, a new model would need to be trained. This is not practically feasible for problems with daily-fluctuating demand, such as the NDP-LS. Therefore, at the price of a slightly lower inference accuracy, we achieve better generalizability by training the model using a dataset that includes different demand profiles. It is shown in all three case studies presented in this section that the single model trained on each network with demand and capacity variations performs well in all NDP-LA and NDP-LS variations studied.

Table 2 and Table 3 summarize the results of the GIN-GA method benchmarked against the SORB algorithm for NDP-LA and NDP-LS based on the Sioux Falls network. As evidenced by the results, GIN-GA achieves better results for NDP-LS for this case study. All solutions obtained for NDP-LA were budget feasible, indicating that the penalty function used in GA objective function to penalize budget-infeasible solutions was effective. For the Sioux Falls network, on average, GIN-GA achieves results with less than 1.5% gap of the optimal results within a minute (which is less than 0.5% of the average total time required for SORB) for NDP-LS. For NDP-LA, this average is about 4.34%, which leads to the average of 2.89% gap for the Sioux Falls network. As indicated by Table 2 and Table 3, the variability within the five replication of GIN-GA for each problem instance is low with the average difference between best and worse results being 0.48% for NDP-LS and 1.62% for NDP-LA, indicating the stability of the framework results. Moreover, narrow 95% confidence intervals for average TTT gaps imply that each stochastic run of GIN-GA is highly likely to yield an average within the indicated interval, which is desirable for the computation time budget provided, even at its upper bound.

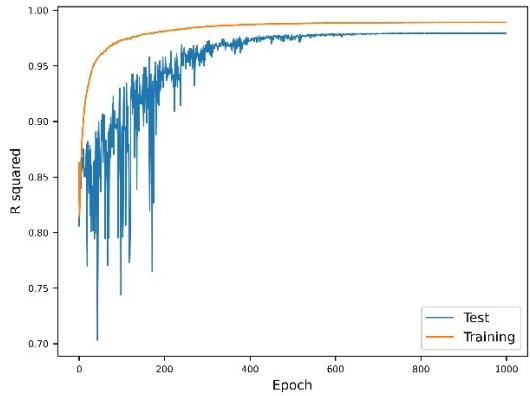 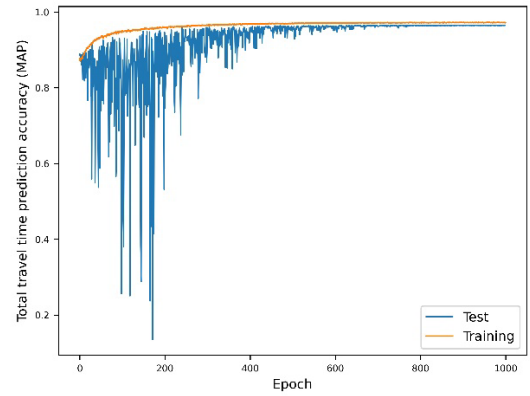

(a) R-squared  (b) TTT prediction accuracy (MAP)

*Figure 2 Summary of GIN training results for Sioux Falls network ((a) R-squared, and (b) prediction accuracy (MAP) of TTT for training and test data).*

*Table 2 TTT and TTT gap comparison between GIN-GA & SORB for NDP-LS on Sioux Falls network (TTT gap values are based on MAPE showing the percentage gap between the TTT value calculated by GIN-GA and SORB with SORB TTT being the reference value. Min, average and max values are based on 5 replications of GIN-GA. 95% confidence intervals are based on the standard error of mean TTT gap values of 5 GIN-GA replications).*

| | | Sioux Falls Network (NDP_LS) | | | | | | | | |
|---|---|---|---|---|---|---|---|---|---|---|
| | | SORB (benchmark) | GIN-GA | | | | | | | |
| Feasible edges | Swapped Lane Capacity | TTT | Min TTT | Average TTT | Max TTT | Min TTT Gap | Average TTT Gap | Max TTT Gap | 95% Confidence Interval (TTT Gap) | |
| 25% | 50% | 7401566.49 | 7526638.97 | 7526638.97 | 7526638.97 | 1.69% | 1.69% | 1.69% | 1.69% | 1.69% |
| 25% | 25% | 7397287.95 | 7467055.63 | 7467055.63 | 7467055.63 | 0.94% | 0.94% | 0.94% | 0.94% | 0.94% |
| 50% | 50% | 7401526.27 | 7526638.97 | 7526638.97 | 7526638.97 | 1.69% | 1.69% | 1.69% | 1.69% | 1.69% |
| 50% | 25% | 7424377.71 | 7519035.93 | 7519035.93 | 7519035.93 | 1.27% | 1.27% | 1.27% | 1.27% | 1.27% |
| 75% | 50% | 7424377.71 | 7529281.50 | 7632973.31 | 7736362.61 | 1.41% | 2.81% | 4.20% | 1.08% | 4.54% |
| 75% | 25% | 7424377.71 | 7441247.86 | 7443222.13 | 7451119.23 | 0.23% | 0.25% | 0.36% | 0.18% | 0.33% |
| Average | | 7412252.30 | 7501649.81 | 7519260.82 | 7537808.56 | 1.21% | 1.44% | 1.69% | 1.14% | 1.74% |

*Table 3 TTT and TTT gap comparison between GIN-GA & SORB for NDP-LA on Sioux Falls network (TTT gap values are based on MAPE showing the percentage gap between the TTT value calculated by GIN-GA and SORB with SORB TTT being the reference value. Min, average and max values are based on 5 replications of GIN-GA. 95% confidence intervals are based on the standard error of mean TTT gap values of 5 GIN-GA replications).*

| | | Sioux Falls Network (NDP_LA) | | | | | | | | |
|---|---|---|---|---|---|---|---|---|---|---|
| | | SORB (benchmark) | GIN-GA | | | | | | | |
| Budget | Added Lane Capacity | TTT | Min TTT | Average TTT | Max TTT | Min TTT Gap | Average TTT Gap | Max TTT Gap | 95% Confidence Interval (TTT Gap) | |
| 25% | 50% | 4857212.76 | 5026025.65 | 5103802.01 | 5230757.49 | 3.48% | 5.08% | 7.69% | 2.93% | 7.22% |
| 25% | 25% | 5726249.39 | 5834988.83 | 5865845.71 | 5925293.25 | 1.90% | 2.44% | 3.48% | 1.67% | 3.20% |
| 50% | 50% | 4383951.83 | 4701676.38 | 4719821.83 | 4738488.78 | 7.25% | 7.66% | 8.09% | 7.22% | 8.10% |
| 50% | 25% | 5309461.76 | 5401115.95 | 5411827.58 | 5422778.30 | 1.73% | 1.93% | 2.13% | 1.70% | 2.16% |
| 75% | 50% | 4357733.23 | 4644231.20 | 4673929.56 | 4689454.57 | 6.57% | 7.26% | 7.61% | 6.75% | 7.76% |
| 75% | 25% | 5284122.32 | 5320690.54 | 5374406.53 | 5405711.38 | 0.69% | 1.71% | 2.30% | 0.85% | 2.57% |
| Average | | 4986455.22 | 5154788.09 | 5191605.54 | 5235413.96 | 3.60% | 4.34% | 5.22% | 3.52% | 5.17% |

### 4.5.2 Case study 2: Eastern-Massachusetts Network

The second case study in this article is based on the Eastern-Massachusetts Network, which is also available at [28]. It is a medium-size network that includes 258 edges, 74 nodes and 5476 OD pairs. Due to higher memory requirements, for the larger case studies, namely, Eastern-Massachusetts and Anaheim networks, we created a dataset with 5,000 solved instances of UE with perturbations in demand and edge capacity. 4,000 of these instances were used for training, 500 for validation, and 500 for testing. Yet, as indicated later in this section, 5000 instances are sufficient for proper training of the GIN model and desirable performance on NDP instances.

The best training and test TTT accuracy for the Eastern-Massachusetts Network were 98.88% and 98.68% respectively, which are remarkably high. The best training and test R-squared for this case study were 99.6% and 99.3% respectively, which indicate good model fit, particularly given how close the training and test statistics are. Eastern-Massachusetts case study training results are summarized in Figure 3.

The trained model is used for inferences within the GIN-GA for all 12 NDP variants for this case study, and the results are shown in Table 4 and Table 5. The average TTT gap achieved with Gin-GA for this case study is 0.80% with average performance on NDP-LA being better (0.48%) than the average on NDP-LS (1.13%). All solutions obtained for NDP-LA were budget feasible. The variability of TTT gaps within replications of GIN-GA and problem instances is remarkably low for this case study with the average difference between the best and the worst results being 0.31% and for NDP-LS and 0.34% for NDP-LA. This low variability along with the narrow confidence intervals for TTT gaps provide compelling evidence for stability and consistency of GIN-GA.

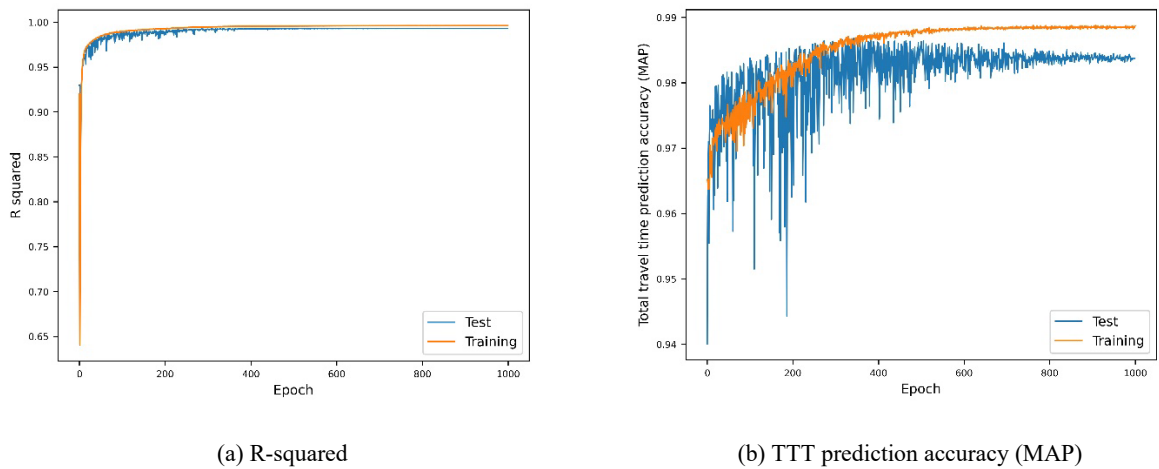

(a) R-squared  (b) TTT prediction accuracy (MAP)

*Figure 3 Summary of GIN training results for Eastern-Massachusetts network ((a) R-squared, and (b) prediction accuracy (MAP) of TTT for training and test data).*

*Table 4 TTT and TTT gap comparison between GIN-GA & SORB for NDP-LS on Eastern-Massachusetts network (TTT gap values are based on MAPE showing the percentage gap between the TTT value calculated by GIN-GA and SORB with SORB TTT being the reference value. Min, average and max values are based on 5 replications of GIN-GA. 95% confidence intervals are based on the standard error of mean TTT gap values of 5 GIN-GA replications).*

| | | SORB (benchmark) | GIN-GA | | | | | | | |
|---|---|---|---|---|---|---|---|---|---|---|
| Feasible edges | Swapped Lane Capacity | TTT | Min TTT | Average TTT | Max TTT | Min TTT Gap | Average TTT Gap | Max TTT Gap | 95% Confidence Interval (TTT Gap) | |
| 25% | 50% | 27330.91 | 27552.40 | 27625.33 | 27684.54 | 0.81% | 1.08% | 1.29% | 0.77% | 1.38% |
| 25% | 25% | 27537.23 | 27782.91 | 27782.91 | 27782.91 | 0.89% | 0.89% | 0.89% | 0.89% | 0.89% |
| 50% | 50% | 27273.24 | 27716.41 | 27746.63 | 27764.36 | 1.62% | 1.74% | 1.80% | 1.64% | 1.83% |
| 50% | 25% | 27462.56 | 27530.04 | 27642.97 | 27716.18 | 0.25% | 0.66% | 0.92% | 0.20% | 1.11% |
| 75% | 50% | 27216.96 | 27745.88 | 27790.11 | 27826.32 | 1.94% | 2.11% | 2.24% | 1.97% | 2.24% |
| 75% | 25% | 27441.20 | 27509.31 | 27534.06 | 27572.40 | 0.25% | 0.34% | 0.48% | 0.23% | 0.44% |
| Average | | 27377.02 | 27639.49 | 27687.00 | 27724.45 | 0.96% | 1.13% | 1.27% | 0.95% | 1.32% |

*Table 5 TTT and TTT gap comparison between GIN-GA & SORB for NDP-LA on Eastern-Massachusetts network (TTT gap values are based on MAPE showing the percentage gap between the TTT value calculated by GIN-GA and SORB with SORB TTT being the reference value. Min, average and max values are based on 5 replications of GIN-GA. 95% confidence intervals are based on the standard error of mean TTT gap values of 5 GIN-GA replications).*

| | | SORB (benchmark) | GIN-GA | | | | | | | |
|---|---|---|---|---|---|---|---|---|---|---|
| Budget | Added Lane Capacity | TTT | Min TTT | Average TTT | Max TTT | Min TTT Gap | Average TTT Gap | Max TTT Gap | 95% Confidence Interval (TTT Gap) | |
| 25% | 50% | 26723.22 | 26869.32 | 26886.14 | 26905.29 | 0.55% | 0.61% | 0.68% | 0.54% | 0.67% |
| 25% | 25% | 27326.66 | 27386.12 | 27450.86 | 27477.16 | 0.22% | 0.45% | 0.55% | 0.29% | 0.62% |
| 50% | 50% | 26720.33 | 26828.82 | 26864.65 | 26888.51 | 0.41% | 0.54% | 0.63% | 0.44% | 0.64% |
| 50% | 25% | 27307.49 | 27347.89 | 27405.29 | 27429.78 | 0.15% | 0.36% | 0.45% | 0.21% | 0.51% |
| 75% | 50% | 26720.29 | 26806.44 | 26897.33 | 27005.04 | 0.32% | 0.66% | 1.07% | 0.28% | 1.05% |
| 75% | 25% | 27307.44 | 27338.40 | 27380.81 | 27417.52 | 0.11% | 0.27% | 0.40% | 0.11% | 0.42% |
| Average | | 27017.57 | 27096.16 | 27147.51 | 27187.22 | 0.29% | 0.48% | 0.63% | 0.31% | 0.65% |

4.5.3    Case study 3: Anaheim Network

Our third case study is a relatively large case based on the Anaheim network, selected from the "transportation networks for research" repository [28]. It entails 914 edges, 416 nodes and 1444 OD pairs. For this case study, similar to the Eastern-Massachusetts Network, we generated a dataset based on 5,000 solved instances of UE using the BFW algorithm, of which, 4,000 were used for training, 500 for validation and 500 for testing. It will be shown in this section that this is a sufficient number of datapoints (i.e., graphs representing instances of UE) for training an effective GIN model for inference within the GIN-GA.

For the Anaheim network, the GIN model used in this study reaches the best training and test accuracy of 97.68% and 97.85% respectively, which indicates outstanding out-of-sample prediction power. The best training and test R-squared achieved for this case study are 97.9% and 97.8% respectively, implying excellent model fit. Figure 4 depicts the evolution of training and test R-squared and TTT prediction accuracy for the Anaheim network.

Regarding the performance of GIN-GA in this case study, as reported in Table 6 and Table 7, the average TTT gaps achieved for NDP-LS and NDP-LA are 0.21% and 1.33% respectively, indicating good performance by GIN-GA, particularly given the one-minute time limit. All solutions obtained for NDP-LA were budget feasible. As for the stability of GIN-GA performance, average gap between the best and the worst-case performance among five replication is 0.1% for NDP-LS and 0.57% for NDP-LA, which shows consistency among replications. The standard errors of TTT gap mean estimation based on the replication performed in this case study are 0.11% and 0.28% for NDP-LS and NDP-LA, respectively, confirming the stability of GIN-GA performance.

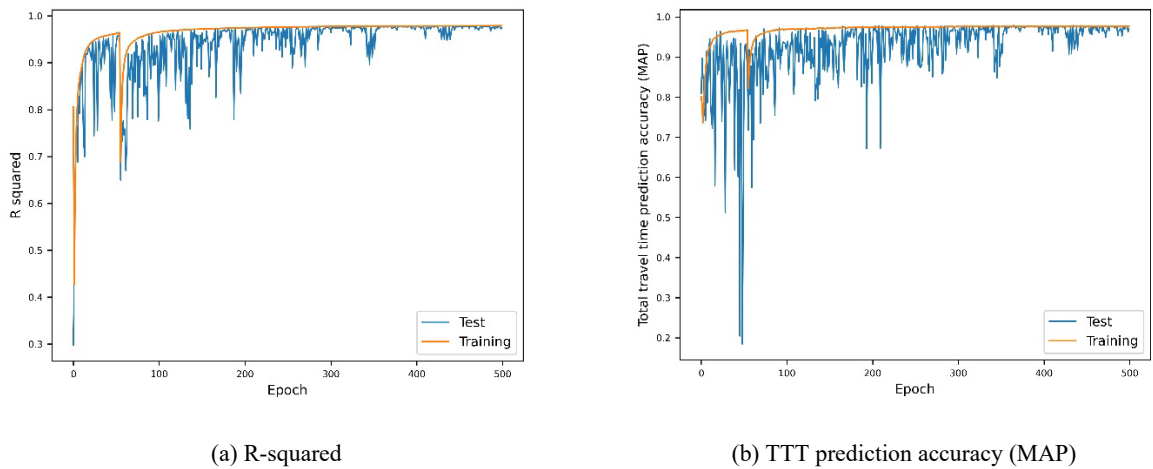

(a) R-squared                                                                  (b) TTT prediction accuracy (MAP)

*Figure 4  Summary of GIN training results for Anaheim network ((a) R-squared, and (b) prediction accuracy (MAP) of TTT for training and test data).*

Table 6  TTT and TTT gap comparison between GIN-GA & SORB for NDP-LS on Anaheim network (TTT gap values are based on MAPE showing the percentage gap between the TTT value calculated by GIN-GA and SORB with SORB TTT being the reference value. Min, average and max values are based on 5 replications of GIN-GA. 95% confidence intervals are based on the standard error of mean TTT gap values of 5 GIN-GA replications).

| | | Anaheim Network (NDP_LS) | | | | | | | | |
|---|---|---|---|---|---|---|---|---|---|---|
| | | SORB (benchmark) | GIN-GA | | | | | | | |
| Feasible edges | Swapped Lane Capacity | TTT | Min TTT | Average TTT | Max TTT | Min TTT Gap | Average TTT Gap | Max TTT Gap | 95% Confidence Interval (TTT Gap) | |
| 25% | 50% | 1331733.95 | 1333162.12 | 1333217.93 | 1333311.46 | 0.11% | 0.11% | 0.12% | 0.10% | 0.12% |
| 25% | 25% | 1332441.45 | 1332849.45 | 1332852.91 | 1332855.22 | 0.03% | 0.03% | 0.03% | 0.03% | 0.03% |
| 50% | 50% | 1331786.88 | 1334619.24 | 1334924.97 | 1335293.73 | 0.21% | 0.24% | 0.26% | 0.21% | 0.27% |
| 50% | 25% | 1331798.79 | 1332401.73 | 1332551.89 | 1333022.10 | 0.05% | 0.06% | 0.09% | 0.03% | 0.08% |
| 75% | 50% | 1331293.53 | 1336253.10 | 1340747.40 | 1342256.17 | 0.37% | 0.71% | 0.82% | 0.47% | 0.95% |
| 75% | 25% | 1331035.85 | 1332725.39 | 1332942.70 | 1333120.56 | 0.13% | 0.14% | 0.16% | 0.13% | 0.16% |
| Average | | 1331681.74 | 1333668.50 | 1334539.63 | 1334976.54 | 0.15% | 0.21% | 0.25% | 0.16% | 0.27% |

Table 7  TTT and TTT gap comparison between GIN-GA & SORB for NDP-LA on Anaheim network (TTT gap values are based on MAPE showing the percentage gap between the TTT value calculated by GIN-GA and SORB with SORB TTT being the reference value. Min, average and max values are based on 5 replications of GIN-GA. 95% confidence intervals are based on the standard error of mean TTT gap values of 5 GIN-GA replications).

| | | Anaheim Network (NDP_LA) | | | | | | | | |
|---|---|---|---|---|---|---|---|---|---|---|
| | | SORB (benchmark) | GIN-GA | | | | | | | |
| Budget | Added Lane Capacity | TTT | Min TTT | Average TTT | Max TTT | Min TTT Gap | Average TTT Gap | Max TTT Gap | 95% Confidence Interval (TTT Gap) | |
| 25% | 50% | 1210379.00 | 1235968.96 | 1240590.09 | 1248174.42 | 2.11% | 2.50% | 3.12% | 1.99% | 3.01% |
| 25% | 25% | 1243757.68 | 1259359.95 | 1262404.13 | 1264404.47 | 1.25% | 1.50% | 1.66% | 1.30% | 1.70% |
| 50% | 50% | 1210114.12 | 1222201.04 | 1227669.29 | 1237950.54 | 1.00% | 1.45% | 2.30% | 0.84% | 2.07% |
| 50% | 25% | 1243919.66 | 1253006.00 | 1255151.78 | 1257226.44 | 0.73% | 0.90% | 1.07% | 0.74% | 1.07% |
| 75% | 50% | 1210119.40 | 1221499.99 | 1222554.52 | 1223978.50 | 0.94% | 1.03% | 1.15% | 0.93% | 1.12% |
| 75% | 25% | 1243926.18 | 1250531.28 | 1251199.22 | 1251880.89 | 0.53% | 0.58% | 0.64% | 0.54% | 0.63% |
| Average | | 1227036.01 | 1240427.87 | 1243261.50 | 1247269.21 | 1.09% | 1.33% | 1.66% | 1.05% | 1.60% |

# 5 Discussion

In this section, we briefly discuss the performance, stability, scalability, and generalizability of GIN-GA results based on the case studies, problems, and variants considered in this article.

## 5.1 Performance

Considering the average TTT gaps obtained by GIN-GA as a surrogate for performance (with the target gap being zero), the average among all NDP-LS problem instances in the three case studies considered was 0.93%, and the average for NDP-LA instances was 2.05%, resulting in an overall average TTT Gap of 1.49%, with gaps as low as 0.03% obtained for some problem instances. Given the one-minute time limit for GIN-GA compared to the four-hour time limit of SORB as benchmark, the results indicate great promise for deployment of GIN-GA in practice. It is noteworthy that transport planners deal with problems such as NDP-LS on a daily basis with slight variations in demand. Therefore, exact solution procedures, such as SORB, are practically infeasible for real-time operations, particularly for large networks due to the exponential growth of computation times of repeatedly solving mixed-integer programing problems. In these settings, finding solutions with one or two percent optimality gaps in one minute with minimal modeling and implementation effort provides excellent value for transport planners and professionals.

## 5.2 Stability

Regarding the stability of GIN-GA performance, small margins between the best- and worst-case performances as well as narrow widths of confidence intervals for mean TTT Gap values are observed throughout case studies and problem instances. The minimum, average and maximum bandwidths for mean confidence intervals (i.e., two times the standard error) were 0%, 0.6%, and 1.72%, respectively, indicating high stability for GIN-GA performance.

## 5.3 Scalability

To assess the scalability of GIN-GA, we evaluate the differences in its performance among different case studies with varied sizes, different NDP variants, and problem instances. Based on the results shown in Table 2, Table 3, Table 4, Table 5, Table 6, and Table 7, there is no systematic difference in the performance of GIN-GA on two NDPs considered in this study, namely, NDP-LA and NDP-LS. Considering the average TTT Gap for each case study as the main indicator, on average GIN-GA performs better in NDP-LA on the Eastern-Massachusetts network, and on NDP-LS on Sioux Falls and Anaheim.

As for the impact of network size on performance, for NDP-LS, the TTT gap decreases on average as the network size increases, indicating a better relative performance compared to SORB for larger networks. And for NDP-LA, the TTT gap is lower for both the mid-sized network (Eastern-Massachusetts) and the large network (Anaheim) compared to the small network (Sioux Falls). Given the constant one-minute computation time limit we enforced on GIN-GA regardless of the network size, one can expect GIN-GA to have a favorable performance on networks larger than the ones considered in this study while there is no evidence in the literature for exact solution procedures of NDPs being capable of dealing with networks larger than Anaheim (900+ edges). It should not go unnoticed that the time required for generating a training dataset with solved UE instances and training the GIN increases with the size of the network.

However, as shown in this study, after a one-time training effort, the trained model can be used for many NDP variants and instances. Moreover, the time can be shortened by reducing the dataset size and/or training epochs.

## 5.4 Generalizability

The GIN model proposed in this study is capable of approximating solutions for a wide range of UE problem instances with variations in demand and capacity. Therefore, it has great potential for use in many NDP variants where the lower level includes variations in demand and/or lane capacity. In this study, we showcase its potential on two NDPs, namely NDP-LS and NDP-LA using different case studies and problem variants. Our results indicated comparable performance on both NDPs, despite the fact that the GIN was trained only once to handle general perturbations on demand and capacity without specific training for a specific NDP. In dealing with all problems considered in this study, GIN approximated numerous unseen UE problem instances, which points to the likelihood of performing well on many other NDPs where the lower-level UE problem can be represented by an attributed graph containing node and edge features.

## 6 Conclusions and future research directions

In this study, we proposed a novel hybrid deep-learning-metaheuristic framework with a bi-level architecture for NDPs. We created datasets of solved instances of UE instances and trained a GIN model to approximate the solutions of UE instances at the lower level and used inferences of the trained GIN in combination with a GA at the upper level to approximate solutions for NDPs. We conducted extensive computational experiments, which entailed solving 30,000 instances of the UE problem on three networks to train the GIN models, as well as generating and approximating 216 instances of NDP using three road networks, two NDPs, six NDP variants for each problem on each network, an exact solution procedure as a benchmark, and five replications of the proposed GIN-GA method. The experiments showed that on average, our proposed framework can provide solutions within 1.5% gap of the optimal results given less than 0.5% of the time used by the exact solution procedure. In addition, we provided evidence for performance stability, scalability, and generalizability of the framework. Our framework can be used in an expert system for infrastructure planning to intelligently determine the optimal infrastructure management decisions. The framework is highly flexible and can be easily adapted to many other decision problems that can be modeled as bi-level problems on graphs. Moreover, this flexibility allows the use of other deep learning models and metaheuristics within the same framework for bi-level problems.

Since this is the first exploratory study on this topic, there are certainly a few shortcomings and areas of improvement that can be further investigated in future research. Regarding the GIN model training, the supervised learning scheme used in this study requires a training dataset, which might occupy large memory space for large networks, need long computation times for generating large datasets, and is dependent on existence of exact solutions for UEs. In addition, inferences made by any deep learning model, including GIN, are always subject to an error margin. This causes inaccuracies in GA, which is not designed to deal with noisy fitness functions evaluations, and thereby pointing out to a need for new metaheuristics.

On the other hand, these shortcomings point to opportunities for interesting future research directions. Therefore, we provide a research agenda below and briefly discuss these opportunities.

Existing metaheuristics are designed to find good solutions in a short amount of time by exploring an exceedingly small fraction of the search space. This is because fitness function evaluations for these algorithms are generally time-consuming. However, inferences made by the GIN model we used in this study for GA fitness evaluations (and GNN inferences in general) are extremely efficient (e.g., each inference made in this study takes less than one millisecond). On the other hand, inferences made by GIN (and GNNs in general) are noisy and never 100% accurate. This points to an opportunity as well as a need for novel heuristics that 1) can cope well with noisy fitness function values, and 2) can use the resulting significantly enlarged computation time provided by GNNs to explore the search space effectively (rather than efficiently). This opens a new avenue for a modern class of metaheuristics that are crafted for use with AI-powered predictors.

While dealing with bi-level programming problems, so far, deep-learning-based frameworks have been used only at the upper level or the lower level. However, they have the potential to be used for both dealing with the upper-level problem as well as the lower-level problem in a bi-level framework. In the previous research agenda item, we mentioned the need for novel heuristics that are tailored to AI-powered predictors for their fitness evaluations. Deep learning frameworks can learn to approximate such heuristics with the mentioned characteristics.

In this study, we selected GIN as our choice of the deep learning model due to its expressive power and showed that it is suitable for our framework. However, other GNNs and deep learning models in general can be used within the framework proposed in this study as well. Future research could investigate the performance of other deep learning models with this framework.

In practice, NDPs should include considerations for multiple user classes (e.g., trucks, manually driven vehicles, automated vehicles) with different behavioral assumptions. These problems are often tackled using heuristics and metaheuristics [37]–[39] since multiple classes with asymmetric travel time interactions make the lower-level problem (UE) non-convex, thereby excluding the exact solution options for the bi-level problem. However, there are solutions available for the multiclass UE problem, which means the GIN introduced in this study can be effectively trained to have comparable performance for NDPs with multiple classes while no exact solution is available for such problems.

We examined the performance of our framework on three networks with 76, 257, and 914 edges, respectively. While for academic purposes, the Anaheim network with 914 edges is considered a rather large network, in practice, NDPs can be defined on road networks with more than 50,000 edges [38], [39]. In such cases, exact solution procedures, such as SORB, are highly unlikely to produce optimal results in practically feasible time frames whereas the framework introduced here is likely to perform well given the magnitude of computation time gains it has provided in this study. However, the computation time and resources required for dealing with large-scale practical NDPs (e.g., with 50,000 edges) makes comprehensive benchmarking studies such as the one presented in this article, which included solving UEs 30,000 times and NDPs 216 times, far from feasible on such networks.